\documentclass[10pt,twocolumn,letterpaper]{article}

\usepackage{cvpr}
\usepackage{times}
\usepackage{epsfig}
\usepackage{graphicx}
\usepackage{amsmath}
\usepackage{amssymb}

\begin{document}

\title{DDAP: Dual-Domain Anti-Personalization against Text-to-Image Diffusion Models}

\author{Jing Yang$^{1}$ \quad\quad Runping Xi$^1$  \quad\quad Yingxin Lai$^2$  \quad\quad Xun Lin$^3$ \quad\quad Zitong Yu$^2$\thanks{Corresponding author}\\
$^1$ Northwestern Polytechnical University \quad\quad
$^2$ Great Bay University \quad\quad
$^3$ Beihang University \\
}
\maketitle
\thispagestyle{empty}

\begin{abstract}
Diffusion-based personalized visual content generation technologies have achieved significant breakthroughs, allowing for the creation of specific objects by just learning from a few reference photos. However, when misused to fabricate fake news or unsettling content targeting individuals, these technologies could cause considerable societal harm. To address this problem, current methods generate adversarial samples by adversarially maximizing the training loss, thereby disrupting the output of any personalized generation model trained with these samples. However, the existing methods fail to achieve effective defense and maintain stealthiness, as they overlook the intrinsic properties of diffusion models. In this paper, we introduce a novel Dual-Domain Anti-Personalization framework (DDAP). Specifically, we have developed Spatial Perturbation Learning (SPL) by exploiting the fixed and perturbation-sensitive nature of the image encoder in personalized generation. Subsequently, we have designed a Frequency Perturbation Learning (FPL) method that utilizes the characteristics of diffusion models in the frequency domain. The SPL disrupts the overall texture of the generated images, while the FPL focuses on image details. By alternating between these two methods, we construct the DDAP framework, effectively harnessing the strengths of both domains. To further enhance the visual quality of the adversarial samples, we design a localization module to accurately capture attentive areas while ensuring the effectiveness of the attack and avoiding unnecessary disturbances in the background. Extensive experiments on facial benchmarks have shown that the proposed DDAP enhances the disruption of personalized generation models while also maintaining high quality in adversarial samples, making it more effective in protecting privacy in practical applications.

\end{abstract}

\section{Introduction}
In recent years, text-to-image generation models have made significant advancements \cite{hoDenoisingDiffusionProbabilistic2020a, songDenoisingDiffusionImplicit2020, rombachHighResolutionImageSynthesis2022a, dhariwalDiffusionModelsBeat2021}, and enabled creation of photorealistic images from a given textual description \cite{rombachHighResolutionImageSynthesis2022a, sahariaPhotorealisticTexttoImageDiffusion2022}. A notable application of these models is personalization, where the aim is to customize text-to-image diffusion models with user-provided subject images \cite{galImageWorthOne2022a, ruizDreamBoothFineTuning2023, kumariMultiConceptCustomizationTexttoImage2023}. Giving just a few reference images, one can generate countless high-quality images containing the specified subject.

\begin{figure}[t]
\begin{center}
\includegraphics[width=0.85\linewidth]{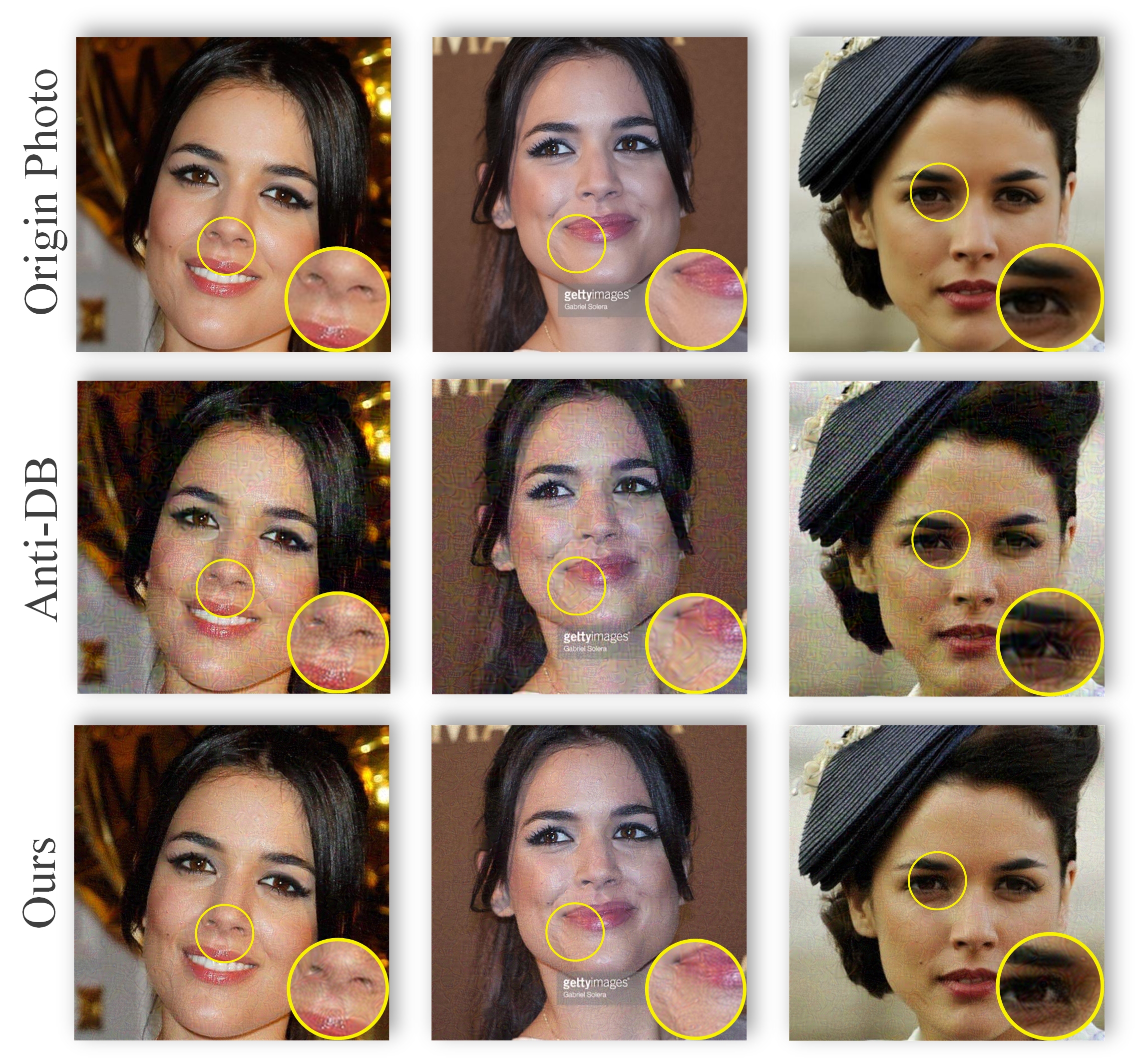}
\end{center}
\vspace{-0.8em}
   \caption{Illustration of adversarial examples generated by the state-of-the-art (SOTA) Anti-DB \cite{vanleAntiDreamBoothProtectingUsers2023}, and our method. The first row displays the original images, the second row features adversarial samples produced by Anti-DB, and the third row showcases adversarial samples from our method. Both methods operate within a noise budget of 12/255.}
\label{fig:protected}
\vspace{-0.8em}
\end{figure}

The advancements in personalized generative models have significantly enhanced the ability to create visual content, presenting a convenient method to produce high-quality and desirable images. However, the development of this technology has also raised concerns about privacy and the authenticity of information. Personalized generative models like DreamBooth \cite{ruizDreamBoothFineTuning2023} and Textual Inversion \cite{galImageWorthOne2022a} enable the production of highly realistic images by fine-tuning specific facial datasets. Such capabilities can be misused to invade privacy, spread misinformation, or harm reputations \cite{wangSecurityPrivacyGenerative2023a}. Additionally, these technologies are sometimes exploited to illegally replicate the artistic styles of others and create unauthorized artwork, violating intellectual property rights \cite{chenPathwayResponsibleAI2023}. Thus, it is imperative to develop effective measures to safeguard users against these malicious applications.

To address these issues, recent proposals\cite{salmanRaisingCostMalicious2023a,liangAdversarialExampleDoes2023a,liangMistImprovedAdversarial2023a,xueEffectiveProtectionDiffusionbased2023,vanleAntiDreamBoothProtectingUsers2023,wangSimACSimpleAntiCustomization2023a} have introduced proactive protections that disrupt fake image generation through adversarial attacks. These methods aim to design perturbations that mislead the personalized diffusion model generation process, thereby impairing its ability to generate. Specifically, Anti-DB\cite{vanleAntiDreamBoothProtectingUsers2023} focuses on disrupting personalized generation methods that utilize fine-tuning, such as DreamBooth\cite{ruizDreamBoothFineTuning2023}. Anti-DB improves attack performance by alternately using a surrogate or fine-tuned fixed model with adversarial perturbations. Although Anti-DB degrades the visual quality of the images, facial features remain recognizable, which is obvious to both human observers and detection algorithms. Additionally, balancing the effectiveness of protection with image quality proves challenging. As shown in Figure~\ref{fig:protected}, despite the same perturbation limits set, the disturbances introduced by Anti-DB are more visually detectable. In practical applications, users tend to choose high-quality images for fine-tuning, and because the disturbances from Anti-DB are prominent, the generated samples are often excluded, weakening its privacy protection in real-world scenarios. Given these limitations, this paper proposes a new solution.

In this paper, we present the Dual-Domain Anti-Personalization framework (DDAP), addressing the challenges discussed earlier through a three-phase approach. Initially, we refined spatial domain attack methods to develop the Spatial Perturbation Learning (SPL) strategy. Using a dual-layer optimization strategy from Anti-DB\cite{vanleAntiDreamBoothProtectingUsers2023} , we effectively targeted fine-tuned personalized models. We introduced a novel loss function to bridge the gap between surrogate and actual models, enhancing defense effectiveness. Inspired by SimAC\cite{wangSimACSimpleAntiCustomization2023a}, which highlighted the diffusion model's sensitivity to image frequency domains, we created the Frequency Perturbation Learning (FPL) method. This method focuses attacks on detailed areas, improving efficiency and reducing irrelevant disturbances. By alternating between SPL and FPL, we established a robust model disruption approach. Additionally, we incorporated Diffusion Attentive Attribution Maps (DAAM)\cite{tangWhatDAAMInterpreting2023} and developed a precision localization module that identifies new concept positions learned by personalized models in images. Targeting these areas optimizes the effectiveness of the disruption and minimizes background disturbances. Our main contributions are summarized as follows:


\begin{itemize}
    \item We introduce the DDAP, a novel anti-personalization framework that integrates our proposed SPL and FPL methods. This framework disrupts the generation process of personalized models by introducing imperceptible perturbations in both the spatial and frequency domains, effectively disturbing the overall texture and fine details of the generated images. 
    \item We design an innovative localization module that can precisely identify the positions of new concepts focused on by personalized models during fine-tuning. By targeting these specific positions for attack, we avoid disturbances in the background and ensure the effectiveness of the attack.
    \item Extensive experiments demonstrate that our method not only enhances the effectiveness of defending personalized models but also maintains the quality of the images, making it better suited for practical applications in protecting user privacy and security.
  \end{itemize}

\section{Related Work}

\subsection{Diffuion Models for Text-to-Image Generation}
Significant advancements have been achieved in text-to-image (T2I) generation, largely thanks to the emergence of large-scale training datasets such as LAION5B \cite{schuhmannLAION5BOpenLargescale2022} and the rapid development in diffusion-based generative models \cite{hoDenoisingDiffusionProbabilistic2020a, dhariwalDiffusionModelsBeat2021,songDenoisingDiffusionImplicit2020}. GLIDE \cite{nicholGLIDEPhotorealisticImage2022} is the first T2I work on diffusion models, which incorporates classifier-free guidance \cite{ho2021classifier} into text-conditioned image synthesis diffusion models, enhancing image quality in terms of photorealism and text-image alignment. DALL-E2 \cite{rameshHierarchicalTextConditionalImage2022a} trains the text encoder jointly with a diffusion prior using paired text-image data. Additionally, several studies \cite{mouT2IAdapterLearningAdapters2024,yangReCoRegionControlledTexttoImage2023,zhangAddingConditionalControl2023} enhance the controllability of T2I synthesis by integrating extra conditional inputs, such as pose, depth, and normal maps. 

Diffusion models typically involve iterative denoising through a large U-Net, making the training computationally intensive. For better trade-off between quality and efficiency, both cascaded diffusion models, as seen in Imagen \cite{sahariaPhotorealisticTexttoImageDiffusion2022}, and Latent Diffusion Models (LDMs \cite{rombachHighResolutionImageSynthesis2022a}) have been introduced. Imagen utilizes cascaded diffusion models in pixel space to produce high-definition videos. Conversely, LDMs compress image data with an autoencoder and apply diffusion models to the resulting latent space, enhancing efficiency. Stable Diffusion, standing out as the first open-source project on latent space, significantly enhances the widespread appllications of T2I generation.

\begin{figure*}
\begin{center}
\includegraphics[width=0.84\linewidth]{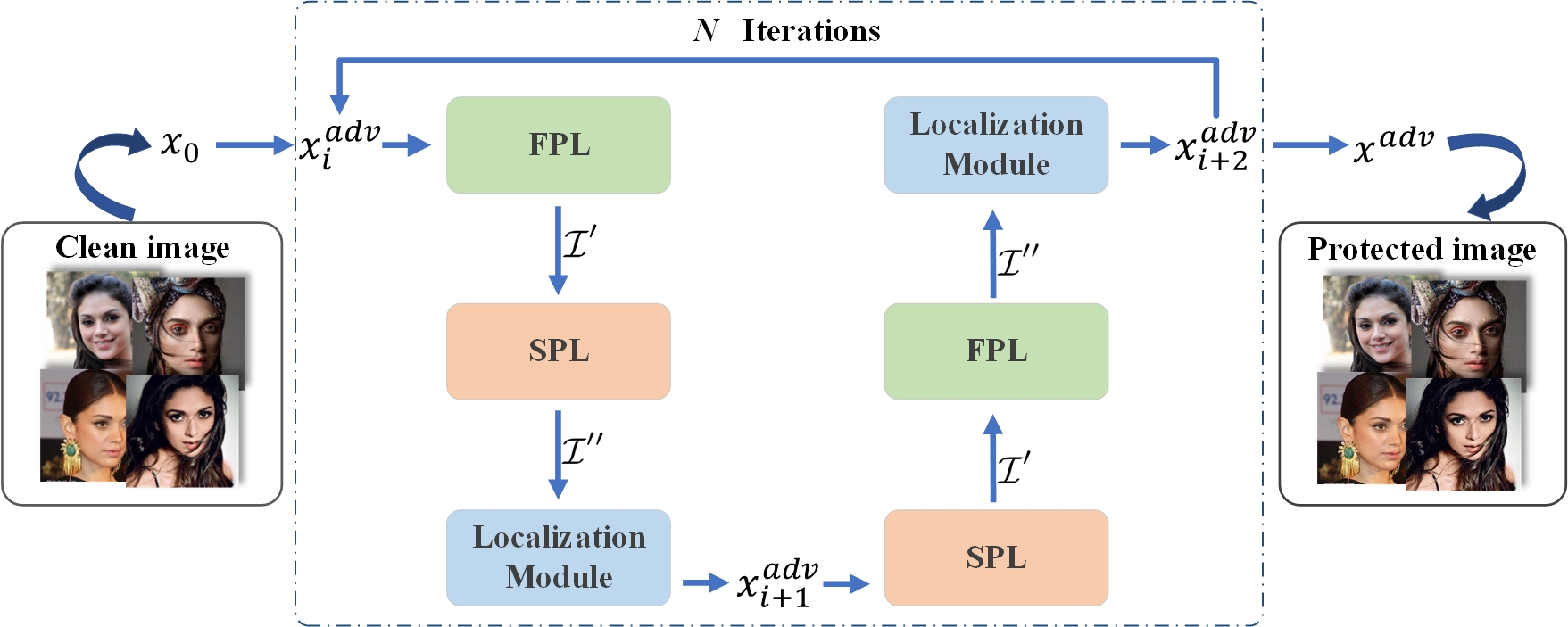}
\end{center}
\vspace{-0.8em}
   \caption{\textbf{Overall architecture}: To integrate the advantages of both perturbation learning methods, we sequentially calculate gradients from the two domains and update the perturbations after filtering through the localization module. We then switch the order of domains. After several iterations, the adversarial samples accumulate gradients from both domains, resulting in improved defense capabilities and maintained stealthiness.}
\label{fig:ddpl}
\end{figure*}

\subsection{Personalization}
Customizing the model to generate a specific person or subject has emerged as a prominent aspect within the field of generative AI. In the era of diffusion model, Textual Inversion \cite{galImageWorthOne2022a} was proposed to optimize a textual embedding of unique identifiers to represent the input concepts. While DreamBooth \cite{ruizDreamBoothFineTuning2023}, standing out as a popular diffusion based methods for personalization, finetunes the the pre-trained Stable Diffusion model with $3 \sim 5$ reference images to connect a less commonly used identifier (e.g., "a sks person") to the new concepts (e.g., a particular person). Aiming to imporve finetuning efficiency, many studies focus on optimizing weight subsets or introducing additional adapters. For instance, CustomDiffusion \cite{kumariMultiConceptCustomizationTexttoImage2023} only optimizes the cross-attention layers in the U-Net. SVDiff \cite{hanSVDiffCompactParameter2023} finetunes the singular values of weights. LoRa \cite{huLoRALowRankAdaptation2021} accelerates the finetuning of personalized models by modifying cross-attention layers based on low-rank adaptation techniques. HyperDreamBooth \cite{ruizHyperDreamBoothHyperNetworksFast2023} represents the input ID images as embeddings further improving the efficiency and speed of the personalization process. Recently, some plug-and-play methods that do not require additional finetuning or training have been proposed. For example, MagicFusion \cite{zhaoMagicFusionBoostingTexttoImage2023} introduces a noise blending method between a pre-trained diffusion model and a T2I personalized model. DreamMatcher \cite{namDreamMatcherAppearanceMatching2024a} enhances the performance of T2I personalized models through semantic matching.

\subsection{Privacy Protection for Diffusion Models}

To prevent private images from being maliciously exploited by Stable Diffusion-based personalization methods, researchers have proposed various techniques to guide models to produce irrelevant outputs. Photoguard \cite{salmanRaisingCostMalicious2023a}, as a pioneering solution, attacks the encoder of the latent diffusion model to prevent it from simulating user-provided images. Building on Photoguard, AdvDM \cite{liangAdversarialExampleDoes2023a} protects user images by generating adversarial samples. Mist \cite{liangMistImprovedAdversarial2023a} combines semantic loss and textural loss to safeguard image privacy. SDS \cite{xueEffectiveProtectionDiffusionbased2023} enhances protection speed and reduces memory usage by introducing Score Distillation Sampling \cite{pooleDreamFusionTextto3DUsing2022}. These methods focus on attacking text-to-image synthesis methods that use fixed diffusion models. In contrast, Anti-DB \cite{vanleAntiDreamBoothProtectingUsers2023} proposes an adversarial strategy to counter fine-tuning-based personalized generation methods like Dreambooth, enhancing protection through alternate training. Building on this, SimAC \cite{wangSimACSimpleAntiCustomization2023a} employs adaptive greedy search to improve protection efficiency. MetaCloak \cite{liu2023toward} uses a meta-learning framework to address the bi-level poisoning issue in Anti-DB, creating perturbations that are both transferable and robust.

Although these methods can mitigate the malicious use of Stable Diffusion for personalized customization to some extent, there are still some limitations that hinder their practical application. The main drawbacks can be summarized in three aspects. Firstly, these methods primarily utilize reconstruction loss for generating adversarial samples, neglecting the unique properties of diffusion models. Secondly, techniques such as the bi-level optimization in Anti-DB \cite{vanleAntiDreamBoothProtectingUsers2023} disrupt the overall texture without effectively erasing the user-provided identity and structural details, compromising privacy protection. Lastly, the approach of perturbing all image pixels results in excessive and irrelevant disturbances, degrading the visual quality particularly in delicate areas like facial features. 

Different from existing works suffering from either weak anti-personalization or poor quality, the proposed DDAP effectively disrupts the generation process of personalized generative models while maintain the balance between perturbation effectiveness and visual quality. 

\section{Methodology}
In this section, we first cover the basics of our task in Section 3.1, including the Latent Diffusion Model (LDM) \cite{hoDenoisingDiffusionProbabilistic2020a}  principles and current adversarial attack methods. In Section 3.2, we discuss our Spatial Perturbation Learning (SPL) that introduces disturbances into pixel space. Section 3.3 details our method for adding disturbances to the frequency domain through Frequency Perturbation Learning (FPL). Section 3.4 describes how we enhance defense by combining SPL and FPL. Lastly, in Section 3.5, we introduce our new localization module. The overall architecture is shown in Figure~\ref{fig:ddpl}.
\subsection{Preliminary}
\textbf{Latent Diffusion Model (LDM)}. The Diffusion Model \cite{hoDenoisingDiffusionProbabilistic2020a} represents a category of generative models. The LDM \cite{rombachHighResolutionImageSynthesis2022a} is a variant that functions within the latent space defined by an autoencoder, as opposed to operating directly in the pixel space.
Specifically, an encoder maps the input image to a latent space $z_0 = \mathcal{E}(x_0)$, and a decoder reconstructs it back to $x_0$.
During the forward process, noise is added at each time step to generate a sequence $\{z_1, z_2,\dots,z_T\}$, where $z_T$ can be considered as a standard Gaussian distribution. 
The backward process trains model $\epsilon_\theta(z_t, t)$ to predict the noise added in $z_t$ to infer $z_{t-1}$. 
During denoising, the training loss is $l_2$ distance, as shown in Eq. \eqref{eq2}.
\begin{eqnarray}
    L_{cond}(\theta, z_0) = E_{z_0,t,P,\epsilon \in \mathcal{N}(0,1)}\|\epsilon - \epsilon_\theta(z_{t+1}, t, P)\|^2_2 . \label{eq2}
\end{eqnarray}
where t is uniformly samples within $\{1,\dots, T\}$, text prompt $P$ is the condition. By iteratively sampling $z_{t-1}$, Gaussian noise $z_T$ is transformed into latent $z_0$. The final image is generated by the decoder $x_0 = \mathcal{D}(z_0)$. 
Stable Diffusion \cite{rombachHighResolutionImageSynthesis2022a} is one of the few open-sourced LDM and is widely used in the community. 
Our research predominantly focused on the experimentation with this model.

\textbf{Adversarial Examples}. 

Unlike adversarial examples in image classification tasks \cite{li2023focus}, which introduce perturbations to cause misclassification, adversarial defense methods for diffusion models aim to prevent personalized generative models from accurately learning image features during training. Additionally, the parameters $\theta$ of the model require finetuning to be obtained. Therefore, the Alternating Surrogate and Perturbation Learning (ASPL) algorithm was proposed in Anti-DB \cite{vanleAntiDreamBoothProtectingUsers2023} to incorporate the training of the surrogate personalization model with the perturbation learning in an alternating manner. The adversarial attack process is shown below:



\begin{figure}[t]
\begin{center}
\includegraphics[width=0.95\linewidth]{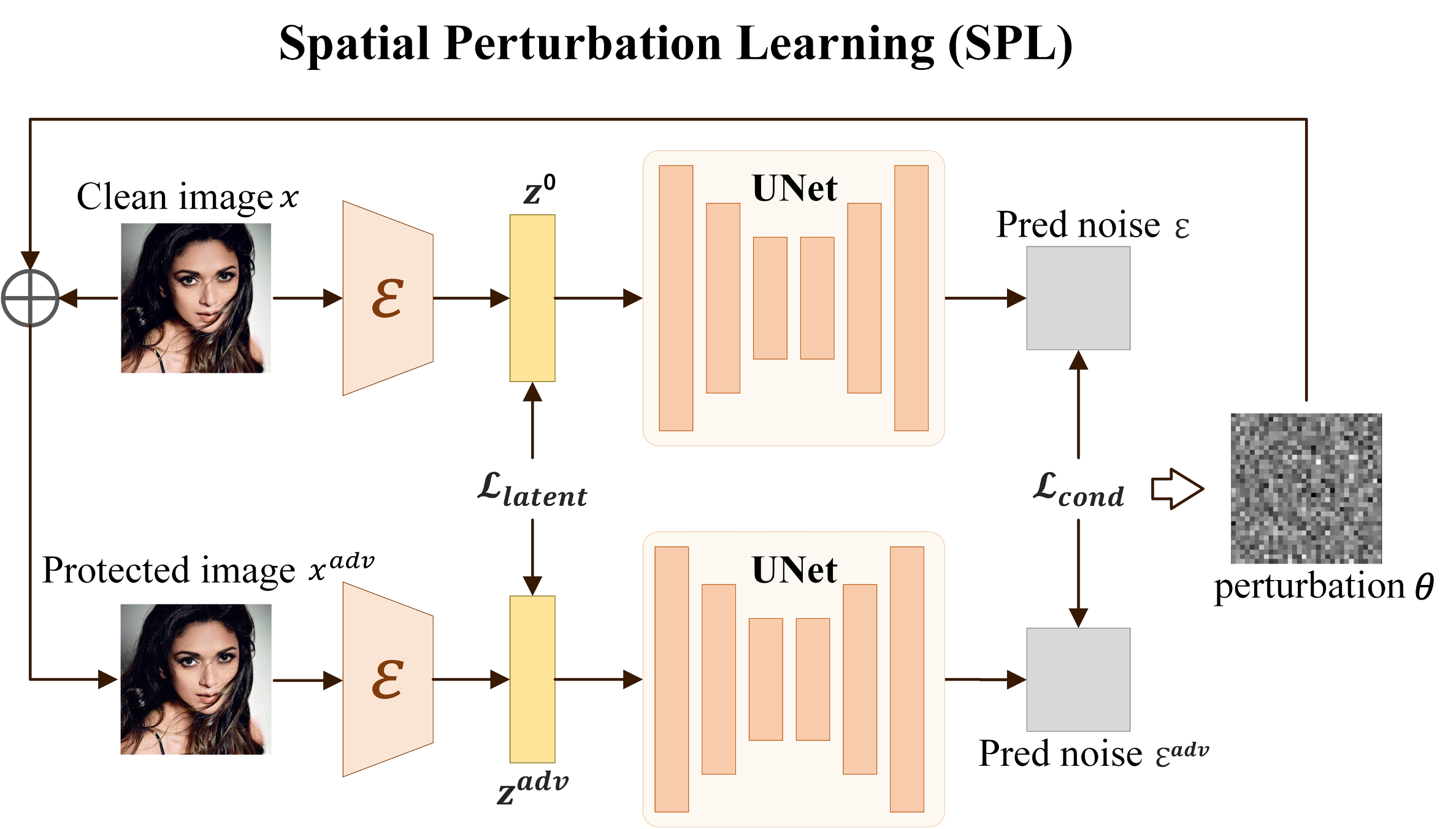}
\end{center}
\vspace{-0.8em}
   \caption{The pipeline of Spatial Perturbation Learning (SPL). This method uses gradients derived from the latent loss of the image and the reconstruction loss of the diffusion model to progressively adjust the generated noise.}
\label{fig:latent}
\vspace{-0.8em}
\end{figure}

    \begin{align}
        \theta' & \gets \theta.clone() , \label{eq4} \\
        \theta' & \gets \underset{\theta'}{\text{arg\,min}}\ \sum_{x \in \mathcal{X}_A} \mathcal{L}_{pn}(\theta', x) ,\\
        \delta_{adv} & \gets \underset{\delta}{\text{arg\,max}}\ \mathcal{L}_{cond}(\theta', x+\delta_{adv}) ,\\
        \theta & \gets \underset{\theta}{\text{arg\,min}}\ \sum \mathcal{L}_{pn}(\theta, x+\delta_{adv}) .
    \end{align}

First, the backbone Stable Diffusion model $\theta$ (Eq. \eqref{eq4}) is copied to $\theta'$. Then, the surrogate personalized generative model $\theta'$ is obtained by minimizing the adversarial loss $\mathcal{L}_{pn}$, which is the loss function for fine-tuning. Subsequently, the adversarial perturbation $\delta$ is obtained by maximizing the conditional loss $\mathcal{L}_{cond}$. Finally, the model parameters $\theta$ are updated by minimizing the fine-tuning loss $\mathcal{L}_{pn}$. This process alternates between the surrogate model and the adversarial perturbation to enhance the effectiveness of the adversarial attack.

\subsection{Spatial Perturbation Learning (SPL)}
Personalized generative models, such as Dreambooth \cite{ruizDreamBoothFineTuning2023}, require thousands of fine-tuning iterations to learn new concepts, typically around 1000 iterations. In contrast, the surrogate model in ASPL \cite{vanleAntiDreamBoothProtectingUsers2023} undergoes only a few fine-tuning iterations, leading to differences between the surrogate and real models. Inspired by \cite{xueEffectiveProtectionDiffusionbased2023}, we propose using a latent loss to target the image encoder module within the LDM. This module remains fixed during fine-tuning and is more vulnerable to perturbations than the denoiser module. The spatial adversarial attack aims to maximize the spatial difference between clean and protected images, causing misidentification in subsequent uses. The process is illustrated in Figure~\ref{fig:latent}, and the objective function includes latent loss and reconstruction loss as follows.
\begin{align}
    \mathcal{L}_{cond} & = E\| \epsilon - \epsilon^{adv} \|^2_2 ,\\    
    \mathcal{L}_{latent} & = E\| z^{0} - z^{adv} \|^2_2 ,\\
    \mathcal{L}_{s} & = \mathcal{L}_{cond} + \xi \mathcal{L}_{latent} , \label{eq9}
\end{align}
where $\epsilon$ and $z$ respectively represent the predicted noise and latent codes generated from images, and $\xi$ represents the weight of the attack on the text encoder.
The objective function of the SPL algorithm is formulated to iteratively update noise for N times to maximize the $\mathcal{L}_{s}$ loss to achieve distinct spitial difference.

\begin{figure}[t]
\begin{center}
\includegraphics[width=0.85\linewidth]{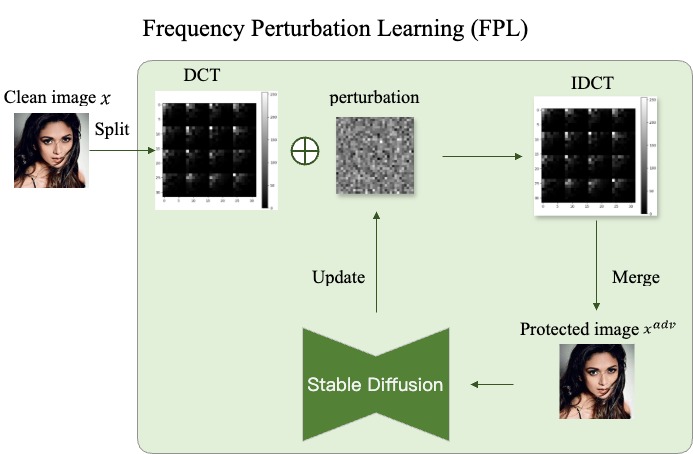}
\end{center}
\vspace{-0.8em}
   \caption{The Frequency Perturbation Learning (FPL) pipeline begins by splitting the input image into blocks and transforming each into the frequency domain using the DCT. Perturbations are added, and then the blocks are converted back to the spatial domain using the IDCT and merged to form the adversarial example. In each iteration, we calculate the adversarial loss and update the perturbations accordingly.}
\label{fig:fre}
\vspace{-0.8em}
\end{figure}

\subsection{Frequency Perturbation Learning (FPL)}
Experiments show that SPL-protected images exhibit significant texture disruption when fine-tuning the DreamBooth model, though details like facial contours remain recognizable. This is because VAE encoders capture high-level features rather than details, and using MSE emphasizes large-area differences at the encoding level. Perturbations introduced by PGD are uniformly distributed, affecting the overall texture. Inspired by SimAC \cite{wangSimACSimpleAntiCustomization2023a}, we found that diffusion models have strong perception in the frequency domain of images. Using smaller denoising time steps significantly enhances the effect of adversarial noise, as high-frequency components become more prominent. Therefore, we propose Frequency Perturbation Learning (FPL), which modifies the image's frequency domain distribution to disrupt high-frequency areas. Compared to previous methods, our approach hides adversarial perturbations within frequency bands, reducing pixel-level redundant noise and making the attack more stealthy while effectively damaging the details of generated images. The FPL process is shown in Figure~\ref{fig:fre}. We summarize the optimization process as follows:

\begin{eqnarray}
    \begin{aligned}
        \arg\max L(\mathcal{D}'(\mathcal{A}(\mathcal{D}(x^{adv}))), \theta, y^{true}),  \\
        s.t. \|\mathcal{D}(x^{adv}) - \mathcal{D}(x^{init})\|_p < \epsilon ,\\
        \|x^{adv} - x^{init}\|_p < \eta,
    \end{aligned}
\end{eqnarray}
where $\mathcal{D}(\cdot)$ denotes the discrete cosine transform (DCT), $\mathcal{D}'$ represents the inverse discrete cosine transform (IDCT), $\mathcal{A}(\cdot)$ is the adjustment module to modify the distribution of the image in the frequency domain. We use $l_p-norm$ to constrain the perturbation in the frequency domain of the original image.

We first transform the image from the spatial domain to the frequency domain using DCT. To balance efficiency and quality, we follow the method in \cite{qianThinkingFrequencyFace2020} and split the image into $K \times K$ blocks before transformation. For each block, the DCT transformation is performed as follows:
\begin{align}
    G(i, j) & = \cos \left[ \frac{(2i+1)u\pi}{2N} \right] \cos \left[ \frac{(2j+1)v\pi}{2M} \right], \\
    \mathcal{D}(u,v) & = c(u) \cdot c(v) \sum_{i=0}^{N-1} \sum_{j=0}^{N-1} x(i, j) G(i, j) ,
\end{align}
where $G(i, j)$ is the product of the cosine components, $x(i,j)$ is the image value at coordinates $(i,j)$, $c(u)$ and $c(v)$ are coefficients, and $N$ is the block size. We then generate the initial adversarial perturbation $\mathcal{P} \sim \mathcal{N}(0,1)$ and add it to the frequency domain information of the image. To optimize the attack, we propose a weight matrix $\mathcal{W}$ that adapts to different images and attack steps. This matrix dynamically adjusts based on the energy distribution of the image's spectrum, assigning greater weight to high-frequency components and less to low-frequency ones, thereby enhancing attack efficiency. The complete Adjustment module is defined as follows:
\begin{eqnarray}
    \mathcal{A}(x^{adv}) = \mathcal{D}(x^{adv}) + \mathcal{W} \odot \mathcal{P} ,
\end{eqnarray}
where $\odot$ denotes the Hadamard product. During the optimization process, $\mathcal{P}_{n+1}$ is updated as follows:
\begin{eqnarray}
    \mathcal{P}_{n+1} = \mathcal{P}_n + \lambda \cdot sign (\nabla_{\mathcal{P}} L(\mathcal{D}'(\mathcal{A}(\mathcal{D}(x^{adv}))), \theta, y^{true})) ,
\end{eqnarray}
where $\lambda$ is the step size, and $n$ is the iteration number. After that, we apply the IDCT to transfer each block back to the spatial domain. Once the maximum number of iterations is reached, we obtain the final adversarial example.

\begin{figure}[t]
\begin{center}
\includegraphics[width=0.85\linewidth]{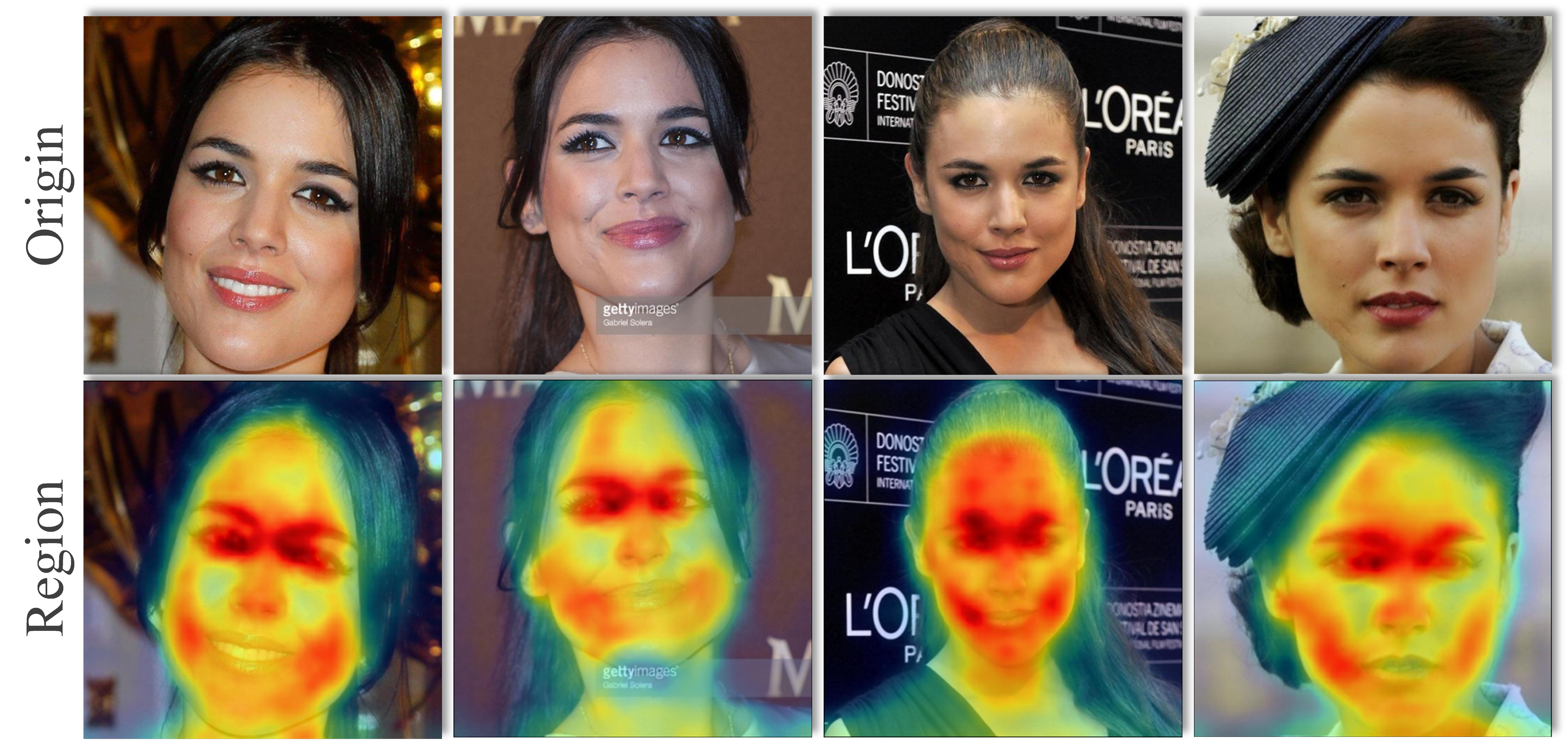}
\end{center}
\vspace{-0.8em}
   \caption{The results of the localization module. The first row displays the original image, while the second row visualizes the areas of focus for new concepts learned by the personalized model during the fine-tuning process.}
\label{fig:attn}
\vspace{-0.8em}
\end{figure}

\subsection{Dual-Domain Perturbation Learning (DDPL)}
Multimodal learning \cite{xuMultimodalLearningTransformers2023} extracts and integrates information from different data sources to enhance learning task performance. Inspired by this, we developed a Dual-Domain Perturbation Learning (DDPL) method that combines information from both the spatial and frequency domains. This approach merges spatial domain structure with frequency domain details, leveraging diffusion model characteristics to improve the stealth and effectiveness of adversarial samples. We will outline our method using formulas.

We denote $\mathcal{A}_{\mathcal{SPL}}$ and $\mathcal{A}_{\mathcal{FPL}}$ as adversarial defends in the spatial domain and frequency domain, respectively. We first update the adversarial perturbation in the frequency domain based on gradient information. The update process is as follows:
\begin{eqnarray}
    \mathcal{I}' =  \mathcal{I} + \gamma_f  \cdot \nabla_{\mathcal{I}} L_{\mathcal{A}_{FPL}}(\mathcal{I}, \theta) ,
\end{eqnarray}
where $\mathcal{I}'$ is frequency domain information, $\gamma_f$ is the step size. We update the adversarial perturbation in the spatial domain based on the updated frequency domain information. The update process is as follows:
\begin{eqnarray}
    \mathcal{I}'' =  \mathcal{I}' + \gamma_l \cdot \nabla_{\mathcal{I}'} L_{\mathcal{A}_{SPL}}(\mathcal{I}', \theta) ,
\end{eqnarray}
where $\mathcal{I}''$ is the updated spatial domain information, and $\gamma_l$ is the step size. The details of the attack in the two domains follow the methods described above. After each iteration, we adjust the sequence between FPL $\mathcal{A}_{\mathcal{FPL}}$ and SPL $\mathcal{A}_{\mathcal{SPL}}$, continuing until a set number of iterations are completed.

\subsection{Localization Module}
To enhance the stealth of adversarial examples and balance the attack's effectiveness with visual quality, we designed a Localization Module that generates attack area masks using Diffusion Attentive Attribution Maps (DAAM) \cite{tangWhatDAAMInterpreting2023}. DAAM employs cross-attention maps to show how textual tokens influence image pixels, guiding image generation in diffusion models. As shown in Figure~\ref{fig:attn}, we obtain a heatmap associated with the new concept by running Denoising Diffusion Implicit Models (DDIM) \cite{songDenoisingDiffusionImplicit2020} inversion on the training image. The attack region mask for adding perturbations is represented as a binary matrix, as shown below:
\begin{eqnarray}
   \mathcal{M}_{s^*}(i, j) = \left\{
    \begin{array}{ll}
      1, &  M_{db}(i, j) \geq \tau \\
      0, & \text{otherwise}
    \end{array}
  \right.
\end{eqnarray}

where $M_{db}$ is the heatmap of the personalized generative model trained by DreamBooth \cite{ruizDreamBoothFineTuning2023} relative to the new concept $S^*$, and $\tau$ is the threshold. By integrating the mask with DDPL, as illustrated in Eq. \eqref{eq18}, we introduce noise into areas related to personalized concepts. 

\begin{eqnarray}
    x^{adv} = x + \delta \odot \mathcal{M}_{s^*}, \label{eq18}
\end{eqnarray}
where $\delta$ is the adversarial perturbation obtained through optimization and $\odot$ denotes the Hadamard product. This approach reduces perturbations in regions irrelevant to the background, making the generated adversarial examples more akin to real images and enhancing the visual effect.

\setlength{\tabcolsep}{3pt} 
\begin{table*}
\begin{center}
\small
\begin{tabular}{|l|c|c|c|c|c|c|c|c|c|c|c|}
\hline
Dateset & Method & \multicolumn{2}{c|}{Quality} & \multicolumn{4}{c|}{“a photo of sks person”} & \multicolumn{4}{c|}{“a dslr portrait of sks person”} \\
\cline{3-12}
 & & PSNR↑ & LPIPS↓ & ISM↓ & FDFR↑ & BRISQUE↑ & SER-FIQ↓ & ISM↓ & FDFR↑ & BRISQUE↑ & SER-FIQ↓ \\
\hline

  & PhotoGurad \cite{salmanRaisingCostMalicious2023a}     & 27.68 & 0.45 & 0.29 & 0.29 & 20.27 & 0.47 & 0.25 & 0.17 & 28.52 & 0.55 \\
\cline{2-12}
VGGFace2 & AdvDM \cite{liangAdversarialExampleDoes2023a}          & 29.44 & 0.42 & 0.32 & 0.63 & 38.51 & 0.21 & 0.30 & 0.68 & 37.58 & 0.35 \\
\cline{2-12}
 & Anti-DB \cite{vanleAntiDreamBoothProtectingUsers2023}        & 28.77 & 0.40  & \textbf{0.21} & 0.76 & 37.33 & 0.22 & 0.23 & 0.86 & 40.92 & 0.26 \\
\cline{2-12}
 & \textbf{DDAP (Ours)} & \textbf{31.93} & \textbf{0.25} & 0.23 & \textbf{0.90} & \textbf{40.19} & \textbf{0.15} & \textbf{0.15} & \textbf{0.88} & \textbf{43.95} & \textbf{0.18} \\
\hline
  & PhotoGurad \cite{salmanRaisingCostMalicious2023a}     & 29.0 & 0.40 & \textbf{0.25} & 0.40 & 19.27 & 0.55 & \textbf{0.20} & 0.29 & 29.52 & 0.59 \\
\cline{2-12}
CelebA-HQ  & AdvDM \cite{liangAdversarialExampleDoes2023a}         & 29.12 & 0.43 & 0.32 & 0.67 & 38.17 & \textbf{0.22} & 0.25 & 0.65 & 37.80 & 0.41 \\
\cline{2-12}
  & Anti-DB \cite{vanleAntiDreamBoothProtectingUsers2023}       & 29.98 & 0.31  & 0.32 & 0.73 & \textbf{38.83} & 0.30 & 0.24 & \textbf{0.67} & 38.96 & \textbf{0.36} \\
\cline{2-12}
  & \textbf{DDAP (Ours)} & \textbf{31.97} & \textbf{0.24} & 0.31 & \textbf{0.75} & 33.87 & 0.25 & 0.23 & \textbf{0.67} & \textbf{39.51} & 0.41 \\
\hline
\end{tabular}
\end{center}
\vspace{-0.8em}
\caption{Comparison with other open-sourced anti-personalization methods on VGGFace2 \cite{caoVGGFace2DatasetRecognising2018} and CelebA-HQ \cite{karrasProgressiveGrowingGANs2018} dateset.}
\label{tab:vgg}
\end{table*}


\begin{figure*}
\begin{center}
\vspace{-0.8em}
\includegraphics[width=0.8\linewidth]{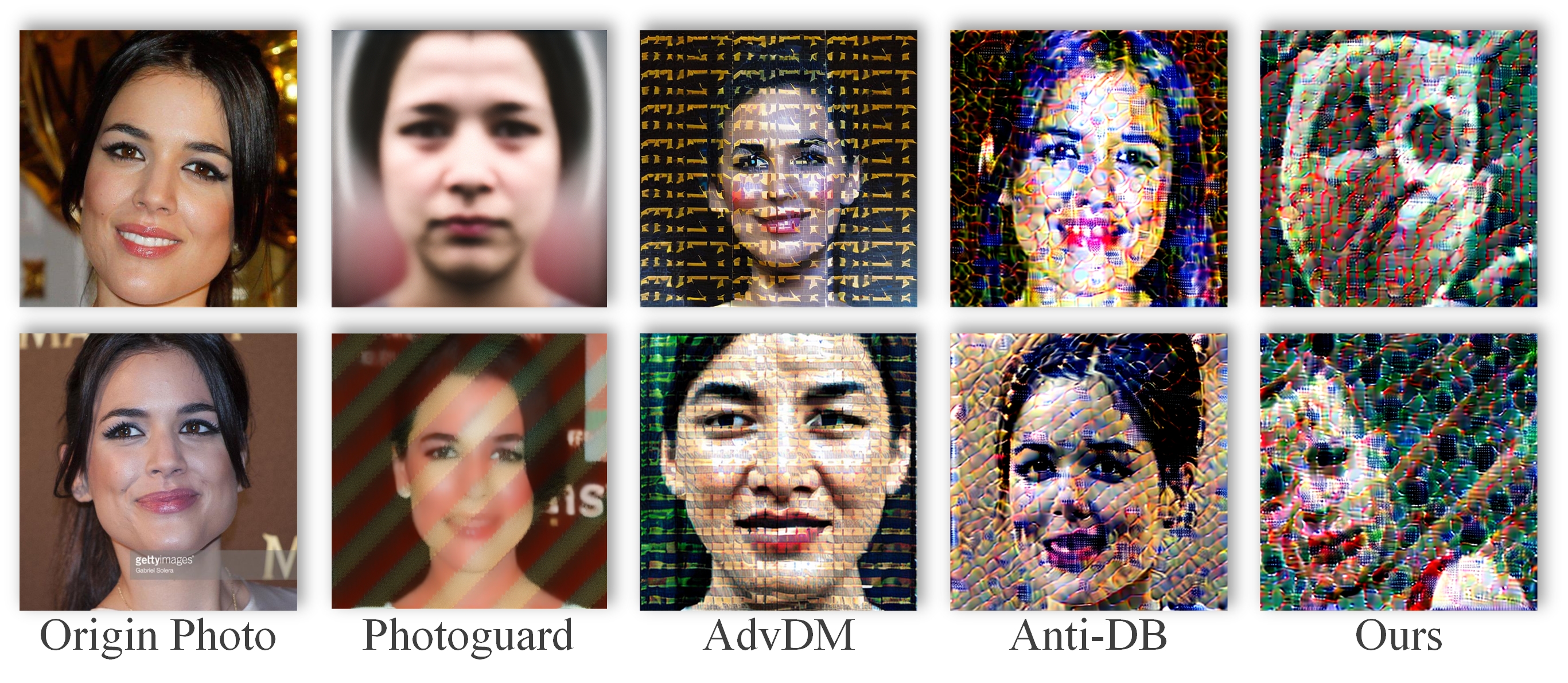}
\end{center}
\vspace{-1.8em}
   \caption{Quantitative results under two prompts. The first row is “a photo of sks person”, the second row is “a dslr portrait of sks person”.}
\label{fig:result}
\vspace{-0.8em}
\end{figure*}

\section{Experiments}
\subsection{Setup} 
\textbf{Datasets.} 
We use two facial datesets for experiments: VGG-Face2 \cite{caoVGGFace2DatasetRecognising2018} and CelebA-HQ \cite{karrasProgressiveGrowingGANs2018} dataset. We randomly selected 50 individuals from these two datasets, consistent with the settings in Anti-DB \cite{vanleAntiDreamBoothProtectingUsers2023}. For each selected individual, we randomly picked 8 images and divided them into two groups: a reference clean image set and a target protection set.

\textbf{Models.} 
Since the most popular open-source LDM \cite{rombachHighResolutionImageSynthesis2022a} is Stable Diffusion, our experiments are mainly conducted on the latest version of Stable Diffusion 2.1. 

\textbf{Baselines.} 
To comprehensively evaluate our method, we compared it with several open-source strategies that prevent T2I generation models from being abused, including Anti-DB \cite{vanleAntiDreamBoothProtectingUsers2023}, PhotoGuard \cite{salmanRaisingCostMalicious2023a}, and AdvDM \cite{liangAdversarialExampleDoes2023a}. We used the default configurations of these methods in the experiments to ensure fairness and consistency in the comparison. All experiments were conducted under the same hardware and software environment to ensure the uniformity of the experimental conditions.

\textbf{Evaluation Metrics.} 
We assess the effectiveness of our method in protecting users from both quantitative and qualitative perspectives. By adding small perturbations to mislead the model's generation process without compromising image quality, we prevent the model from learning user-specific image details. Our evaluation metrics include assessing the quality of the input image after perturbation, evaluating the generated image quality, and measuring the similarity between the generated and input images.
To evaluate the perturbed image quality, we use LPIPS for perceptual similarity and PSNR for signal-to-noise ratio. We employ Retinaface\cite{dengRetinaFaceSingleShotMultiLevel2020} to determine the Face Detection Failure Rate (FDFR) in generated images, indicating how many faces remain undetected. For detected faces, ArcFace\cite{dengArcFaceAdditiveAngular2019} calculates the average Identity Score Matching (ISM) with the user's clean image set. For evaluating the generated image quality, BRISQUE\cite{mittalNoReferenceImageQuality2012} assesses naturalness, while SER-FIQ\cite{terhorstSERFIQUnsupervisedEstimation2020} evaluates authenticity.

\textbf{Implementation Details.}
We have set the same noise budget for all methods, which is $\eta = 12/255$. Additionally, the optimization steps and step sizes are aligned with those specified in each benchmark test. Our training regimen spans 50 epochs, each consisting of three steps to train the surrogate model and nine steps to optimize the adversarial noise. The default step lengths for adding noise to the spatial and frequencies are set at 0.005 and 0.1, respectively. For attacks targeting the encoder module, we have set $\lambda = 5e-3$. The default configuration involves using Stable Diffusion v2.1 combined with DreamBooth, where we fine-tuned the text encoder and U-Net model with a learning rate of $5 \times 10^{-7}$  and a batch size of 2, over 1000 iterations. For each prompt, we generated 16 images and used them to compute the metrics.

\subsection{Comparison with Baseline Methods}
Here are some results shown in Figure~\ref{fig:result}. It is evident that DDPM combined with Anti-DB achieves strong image disruption effects, providing optimal privacy protection for input portraits. In contrast, PhotoGuard operates attacks in the latent space, often generating images that exhibit patterns similar to the target latent space. Additionally, both AdvDM and Anti-DB optimize noise using DM's training loss, resulting in similar outcomes where the overall texture of generated images is disrupted. Despite the decrease in image quality, AdvDM and Anti-DreamBooth retain many details from the user input images, potentially compromising user privacy. In comparison, our method leverages both spatial and frequency domain characteristics and achieves superior facial privacy protection compared to all other methods under the same noise budget.

\textbf{Quantitative Results.} 
As shown in Table~\ref{tab:vgg}, our proposed method outperforms other baselines in most metrics. Particularly, on the crucial metrics FDFR and ISM, our method achieves a 14\% improvement in FDFR on the VGGFace2 dataset compared to the SOTA method. It also maintains good efficiency on the CelebA-HQ dataset. Furthermore, quantitative results from PSNR and LPIPS demonstrate that our method effectively preserves image quality while retaining strong defense capabilities. Overall, our method performs well in protecting user privacy. However, due to the higher quality of images in the CelebA-HQ dataset, the protection effectiveness under the same noise budget is not as strong as observed in the results from the VGGFace2 dataset.

\subsection{Ablation Study}
We conduct a series of ablation studies on the proposed attack method. For simplicity, we only use VGG-Face2 dataset for the ablation study. More ablation study are listed in supplementary materials.

\setlength{\tabcolsep}{3pt} 
\begin{table}
\small
\begin{center}
\small
\begin{tabular}{|l|c|c|c|c|c|c|}
\hline
Method & \multicolumn{2}{c|}{Quality} & \multicolumn{4}{c|}{“a photo of sks person”} \\
\cline{2-7}
& PSNR↑ & LPIPS↓ & ISM↓ & FDFR↑ & BRISQUE↑ & SER-FIQ↓ \\
\hline
SPL & 29.09 & 0.35 & 0.24 & 0.92 & 38.88 & 0.20 \\
\hline
FPL & 32.01 & 0.21 & 0.39 & 0.60 & 20.62 & 0.37 \\
\hline
DDAP & 31.93 & 0.25 & 0.23 & 0.90 & 40.19 & 0.14 \\
\hline
\end{tabular}
\end{center}
\vspace{-0.8em}
\caption{Ablation results on VGGFace2 \cite{caoVGGFace2DatasetRecognising2018}.}
\label{tab:variants}
\end{table}

\textbf{Variants of Our Method.}
We have designed several variants to assess the efficacy of our approach and to explore the impact across different domains. Specifically, we have introduced the following variants: 1) SPL used exclusively for adversarial attacks in the spatial domain; 2) FPL employed solely for adversarial attacks in the frequency domain; 3) DDAP applied for adversarial attacks across both spatial and frequency domains. The average scores for these variants are presented in Table~\ref{tab:variants}. It is evident that the approach focusing solely on the spatial domain exhibits inferior performance in terms of adversarial sample quality, whereas the approach limited to the frequency domain shows weaker defensive capabilities. This underscores the effectiveness of our method's integration across both domains.

\setlength{\tabcolsep}{3pt} 
\begin{table}
\small
\begin{center}
\small
\begin{tabular}{|l|c|c|c|c|c|c|}
\hline
Method & \multicolumn{2}{c|}{Quality} & \multicolumn{4}{c|}{“a photo of sks person”} \\
\cline{2-7}
& PSNR↑ & LPIPS↓ & ISM↓ & FDFR↑ & BRISQUE↑ & SER-FIQ↓ \\
\hline
w/o  & 30.10     & 0.29             & 0.23     & 0.83    & 36.88    & 0.12     \\
\hline
w & 31.93     & 0.25    & 0.23     & 0.90    & 40.19    & 0.15 \\
\hline
\end{tabular}
\end{center}
\vspace{-0.8em}
\caption{The influence of Localization Module on VGGFace2 \cite{caoVGGFace2DatasetRecognising2018}.}
\label{tab:local}
\end{table}

\setlength{\tabcolsep}{3pt} 
\begin{table}
\small
\begin{center}
\small
\begin{tabular}{|l|c|c|c|c|}
\hline
DreamBooth prompt & \multicolumn{4}{c|}{“a photo of sks person”} \\
\cline{2-5}
& ISM↓ & FDFR↑ & BRISQUE↑ & SER-FIQ↓ \\
\hline
“sks” → “sks”      & 0.23     & 0.90    & 40.19    & 0.15     \\
\hline
“sks” → “t@t”   & 0.25     & 0.71    & 39.95    & 0.52     \\
\hline
\end{tabular}
\end{center}
\vspace{-0.8em}
\caption{Prompt mismatch which during training and testing on VGGFace2 \cite{caoVGGFace2DatasetRecognising2018}. The training prompt is “a photo of sks person” and the inference prompt is combined with the rare identifiers “sks” or “t@t”.}
\label{tab:prompt}
\end{table}

\textbf{Effectiveness of Localization Module.}  In order to verify the effectiveness of the proposed Localization Module in enhancing the stealthiness of adversarial examples, we conducted an ablation study. Specifically, we removed the Localization Module from our method and conducted experiments on the VGG-Face2 dataset. As shown in Table~\ref{tab:local}, the method without the Localization Module performs poorly in PSNR and LPIPS, indicating that the Localization Module is effective in enhancing the stealthiness of adversarial examples.

\setlength{\tabcolsep}{3pt} 
\begin{table}
\small
\begin{center}
\small
\begin{tabular}{|l|c|c|c|c|}
\hline
Model & \multicolumn{4}{c|}{“a photo of sks person”} \\
\cline{2-5}
& ISM↓ & FDFR↑ & BRISQUE↑ & SER-FIQ↓ \\
\hline
v2.1 → v2.1      & 0.23     & 0.90    & 40.19    & 0.15     \\
\hline
v2.1 → v1.5   & 0.01        &99.75    & 55.63    & 0.02    \\
\hline
v1.5 → v2.1   & 0.27     & 0.77    & 39.03    & 0.44  \\
\hline
\end{tabular}
\end{center}
\vspace{-0.8em}
\caption{Model versions mismatch during training and testing on VGGFace2\cite{caoVGGFace2DatasetRecognising2018}. The training prompt is “a photo of sks person”.}
\label{tab:model}
\end{table}
\subsection{Black-Box Attack}
In this subsection, we investigate the performance of our method in black-box attacks, i.e., whether the proposed defense is still effective when some components are unknown. All our experiments will use Stable Diffusion v2.1 and be conducted on the VGGFace2 dataset. We divide black-box attacks into two cases: 1) prompt mismatch, where the attacker uses different prompts; 2) model mismatch, where the attacker uses a different Latent diffusion model. 

\textbf{Prompt Mismatch.}
When attackers employ stable diffusion for concept customization, the prompts they use during the addition of noise may vary from our initial assumptions. Therefore, we utilize the prompt "a photo of sks person" in our perturbation learning phase and replace the unique identifier "sks" with "t@t" during the fine-tuning of the DreamBooth model. The data shown in Table~\ref{tab:prompt} reveal a performance decline under these conditions, but we notice that the ISM, representing identity similarity remains consistent.

\textbf{Model Mismatch.}
The models used to add adversarial noise may also mismatch with the model fine-tuned by Dreambooth. We examine the effectiveness of adversarial noise learned on stable diffusion v2.1 against customization based on stable diffusion v1.5, or vice versa in Table~\ref{tab:model}. We observed a decline in performance when training on SD v1.5 and testing on SD v2.1. We suspect this may be due to differences between the v2.1 and v1.5 datasets, resulting in reduced perturbation performance on person.


\section{Conclusion}
In this paper, we introduce a novel Dual-Domain Anti-Personalization approach to safeguard user privacy in text-to-image generation models. This method synergistically combines the strengths of both the spatial and frequency domains, substantially improving the confidentiality and effectiveness of the adversarial samples. Additionally, we develop a Localization Module that further enhances the balance between attack efficiency and visual quality. Extensive experiments validate the efficacy of our approach in protecting user privacy. Future work will focus on enhancing transmission defenses against unauthorized personalization across various scenarios.

\textbf{Acknowledgement.}  This work was supported by National Natural Science Foundation of China under Grant 62306061, and Guangdong Basic and Applied Basic Research Foundation (Grant No. 2023A1515140037).

{\small
\bibliographystyle{ieee}
\bibliography{main}
}

\end{document}